\documentclass[a4paper]{article}

\usepackage{INTERSPEECH2022}

\title{Speech Segmentation Optimization using Segmented Bilingual Speech Corpus for End-to-end Speech Translation}
\name{Ryo Fukuda, Katsuhito Sudoh, Satoshi Nakamura}
\address{Nara Institute of Science and Technology, Japan}
\email{\{fukuda.ryo.fo3, sudoh, s-nakamura\}@is.naist.jp}

\usepackage{tabularx}

\newcommand{\argmax}{\mathop{\rm arg~max}\limits}

\begin{document}

\maketitle
 
\begin{abstract}
Speech segmentation, which splits long speech into short segments, is essential for speech translation (ST).
Popular VAD tools like WebRTC VAD\footnote{https://github.com/wiseman/py-webrtcvad} have generally relied on pause-based segmentation.
Unfortunately, pauses in speech do not necessarily match sentence boundaries, and sentences can be connected by a very short pause that is difficult to detect by VAD.
In this study, we propose a speech segmentation method using a binary classification model trained using a segmented bilingual speech corpus.
We also propose a hybrid method that combines VAD and the above speech segmentation method.
Experimental results reveal that the proposed method is more suitable for cascade and end-to-end ST systems than conventional segmentation methods.
The hybrid approach further improves the translation performance.
\end{abstract}
\noindent\textbf{Index Terms}: speech translation, segmentation, bilingual speech corpus

\section{Introduction}
\label{sec:introduction}
The segmentation of continuous speech is one process required for speech translation (ST).
In text-to-text machine translation (MT), an input text is usually segmented into sentences using punctuation marks as boundaries.
However, such explicit boundaries are unavailable in ST.
Pre-segmented speech in a bilingual speech corpus can be used when training an ST model, but not in realistic scenarios.
Since existing ST systems cannot directly translate long continuous speech, automatic speech segmentation is needed.
\par
Pause-based segmentation using voice activity detection (VAD) is commonly used for segmenting speech for preprocessing automatic speech recognition (ASR) and ST, even though pauses do not necessarily coincide with semantic boundaries.
Over-segmentation, in which a silence interval fragments a sentence, and under-segmentation, in which multiple sentences are included in one segment while ignoring a short pause, are problems that reduce the performances of ASR and ST \cite{wan2021segmenting}.
\par
In this study, we propose a novel speech segmentation method\footnote{
There is an independent work by Tsiamas et al., (2022) (https://arxiv.org/abs/2202.04774).} based on the segment boundaries in a bilingual speech corpus.
ST corpora usually include bilingual text segment pairs with corresponding source language speech segments.
If we can similarly segment input speech to such ST corpora, we might successfully bridge the gap between training and inference.
Our proposed method uses a Transformer encoder \cite{vaswani2017attention} that was trained to predict segment boundaries as a frame-level sequence labeling task for speech inputs.
\par
The proposed method is applicable to cascade and end-to-end STs because it directly segments speech.
Cascade ST is a traditional ST system that consists of an ASR model and a text-to-text MT model.
An end-to-end ST uses a single model to directly translate source language speech into target language text.
We limit our scope to a conventional end-to-end ST, which treats segmentation as an independent process from ST.
Integrating a segmentation function into end-to-end ST is future work.
\par
We conducted experiments with cascade and end-to-end STs on MuST-C \cite{di2019must} for English-German and English-Japanese.
In the English-German experiments, the proposed method achieved improvements of -9.6 WER and 3.2 BLEU for cascade ST and 2.7 BLEU for end-to-end ST compared to pause-based segmentation.
In the English-Japanese experiments, the improvements were -9.6 WER and 0.5 BLEU for cascade ST and 0.6 BLEU for end-to-end ST.
A hybrid method with the proposed segmentation model and VAD further improved both language pairs.

\section{Related work}
Early studies on segmentation for ST considered modeling with the Markov decision process \cite{mansour2010morphtagger, sinclair2014semi}, conditional random fields \cite{nguyen-vogel-2008-context, lu2010better}, and support vector machines \cite{diab-etal-2004-automatic, sadat-habash-2006-combination, matusov-etal-2007, rangarajan-sridhar-etal-2013-segmentation}.
They focused on cascade ST systems that consist of an ASR model and a statistical machine translation (SMT) model, which were superseded by newer ST systems based on neural machine translation (NMT).
\par
In recent studies, many speech segmentation methods based on VAD have been proposed for ST.
Gaido et al. \cite{DBLP:journals/corr/abs-2104-11710} and Inaguma et al. \cite{inaguma-etal-2021-espnet} used the heuristic concatenation of VAD segments up to a fixed length to address the over-segmentation problem.
Gállego et al. \cite{gallego2021end} used a pre-trained ASR model called wav2vec 2.0 \cite{baevski2020wav2vec} for silence detection.
Yoshimura et al. \cite{yoshimura2020end} used an RNN-based ASR model to consider consecutive blank symbols (``\_") as a segment boundary in decoding using connectionist temporal classification (CTC).
Such CTC-based speech segmentation has an advantage; it is easier to intuitively control segment lengths than with a conventional VAD
because the number of consecutive blank symbols that is considered a boundary can be adjusted as hyperparameters.
However, these methods often split audio at inappropriate boundaries for ST because they mainly segment speech based on long pauses.
\par
Re-segmentation using ASR transcripts is widely used in cascade STs.
Improvements in MT performance have been reported by re-segmenting transcriptions to sentence units using punctuation restoration \cite{lu2010better, rangarajan-sridhar-etal-2013-segmentation, cho2015punctuation, ha2015kit, cho17_interspeech} and language models \cite{stolcke1996automatic, wang2016efficient}.
Unfortunately, they are difficult to use in end-to-end ST and cannot prevent ASR errors due to pause-based segmentation.
We discuss ASR degradation due to speech segmentation with VAD in section \ref{subsec:results-baselines}.
\par
Finally, we refer to speech segmentation methods based on segmented speech corpora that are more relevant to our study.
Wan et al. \cite{wan2021segmenting} introduced a re-segmentation model for modifying the segment boundaries of ASR output using movie and TV subtitle corpora.
Wang et al. \cite{wang2019online} and Iranzo-Sánchez et al. \cite{iranzo-sanchez-etal-2020-direct} proposed an RNN-based text segmentation model using the segment boundaries of a bilingual speech corpus.
Unlike these methods that require ASR transcripts, our proposed method directly segments speech and can be used in an end-to-end ST as well as in a cascade ST.
\section{Proposed method} \label{sec:method} Our method defines speech segmentation as a frame-level sequence labeling task of acoustic features.
This section describes the following details of it: the process that extracts training data for the speech segmentation task from the speech translation corpus (\ref{subsec:data}), the architecture of the proposed model (\ref{subsec:model}), and the training (\ref{subsec:training}) and inference algorithms (\ref{subsec:inference}).

\subsection{Data extraction} \label{subsec:data}
\begin{figure}[t]
\centering
\includegraphics[keepaspectratio, width=7.7cm]{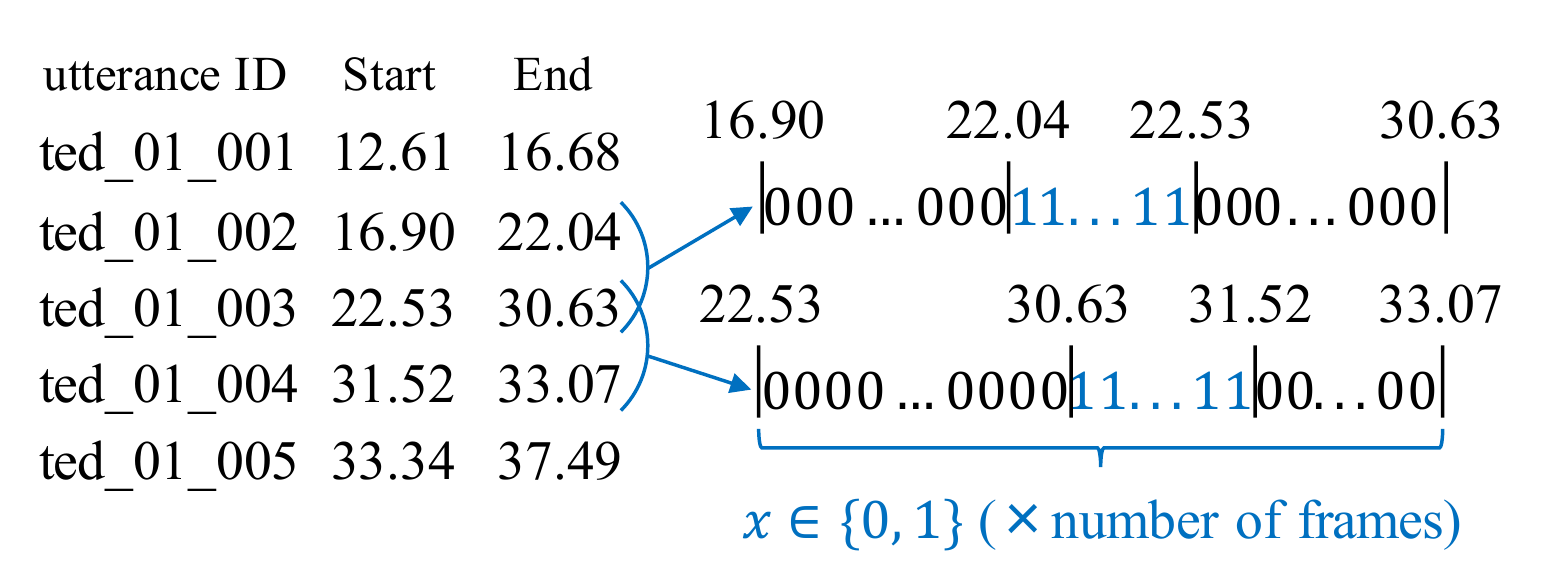}
\caption{Data extraction examples from MuST-C: Labels 0 and 1 are assigned to each FBANK frame based on each segment's starting and ending times.}
\label{fig:sample}
\end{figure}

We use a bilingual speech corpus that includes speech segments aligned to sentence-like unit text as training data for speech segmentation from continuous speech into units suitable for translation.
Data extraction examples are shown in Fig. \ref{fig:sample}.
Two consecutive segments are concatenated and assigned a label, $x\in\{0, 1\}$, representing that the corresponding frame is inside (0) and outside (1) of the utterance.

\setlength\textfloatsep{15pt}

\subsection{Segmentation model} \label{subsec:model}
\begin{figure}[t]
\centering
\includegraphics[keepaspectratio, width=5.9cm]{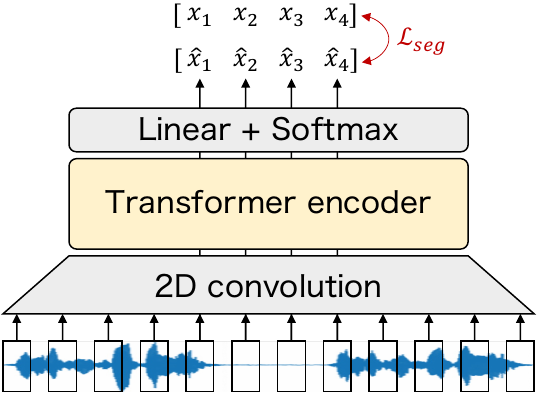}
\caption{Transformer encoder-based speech segmentation model}
\label{fig:model}
\end{figure}

The architecture of the proposed speech segmentation model is illustrated in Fig. \ref{fig:model}.
It consists of a 2D convolution layer, Transformer encoder layers, and an output layer (Linear+Softmax), which outputs label probability $\hat{x}_n\in R^2$ at the $n$-th frame of the convolution layer.
The convolution layer reduces the sequence length by a quarter to handle long speech sequences.
Downsampling is applied to teacher label $x$ to align its length with the model’s output.
Here, labels are simply extracted at regular time intervals calculated from the input-output length ratio.

\subsection{Training} \label{subsec:training}
The model is trained using the data extracted in section \ref{subsec:data} and
learns to minimize cross-entropy loss $\mathcal{L}_{seg}(\hat{x},x)$ between prediction $\hat{x}$ and label $x$:

\begin{equation}
\label{eq:loss}
\begin{split}
\mathcal{L}_{seg}(\hat{x},x):=-\sum_{n=1}^N\left\{
w_s\log\frac{\exp(\hat{x}_{n,1})}{\exp(\hat{x}_{n,0}+\hat{x}_{n,1})}x_{n,1}\right. \\
\left.+~(1-w_s)\log\frac{\exp(\hat{x}_{n,0})}{\exp(\hat{x}_{n,0}+\hat{x}_{n,1})}x_{n,0}
\right\}
\end{split},
\end{equation}

where $w_s$ is a hyperparameter that adjusts the weights of the unbalanced labels.
Since most of the labels are $0$ (within utterances), we increase the weight on the loss of label $1$ (outside utterances).
We set $w_s=0.9$ based on the best loss of the validation data in a preliminary experiment.

\subsection{Inference} \label{subsec:inference}
During inference, speech is segmented at a fixed-length $T$ and input independently into the segmentation model.
Fixed-length segments are then resegmented according to the labels predicted by the segmentation model.
The segmentation model simply selects label $l_n \in \{0,1\}$ with the highest probability at each time $n$:
\begin{equation}
    \label{eq:single}
    l_n := \argmax(x_n).   
\end{equation}

\subsubsection{Hybrid method} \label{subsubsec:hybrid}
In addition, to make the predictions more appropriate, we combined the model predictions calculated in Eq. (\ref{eq:single}) with the results of the VAD tool:
\begin{equation}
    \label{eq:hybrid}
    l_n := \left\{
    \begin{array}{ll}
    \argmax(x_n) \land \mathrm{vad}_n & (len(n-1) < maxlen) \\
    \argmax(x_n) \lor \mathrm{vad}_n & (len(n-1) \geq maxlen), \\
    \end{array}
    \right.
\end{equation}
where $\mathrm{vad}_n \in \{0,1\}$ is the VAD output at time $n$, corresponding to the active and inactive frames.
Segment length $len(n) $ at time $n$ is denoted as follows:

\begin{equation}
    len(n) := \left\{
    \begin{array}{lr}
    len(n-1) + 1&(l_n = 1) \\
    0&(l_n = 0) \\
    \end{array},
    \right.
\end{equation}
where $len(0) := 0$.
Eq. (\ref{eq:hybrid}) implies that a frame can only be included in a segment boundary when segmentation model \textbf{and} VAD agree.
This agreement constraint may result in very long segments, so we relax it not to require the agreement once the segment length exceeds $maxlen$.
In that case, a frame can be included in a segment boundary if either the segmentation model \textbf{or} the VAD allows it.

\section{Experimental setup}
\subsection{Tasks} \label{subsec:task}
We conducted experiments with English to German and English to Japanese ST.
We used MuST-C v1 for the English-German and v2 for the English-Japanese experiments.
Each dataset consisted of triplets of segmented English speech, transcripts, and target language translations.
The English-German and English-Japanese datasets contained about 230k and 330k segments.
As acoustic features, we used 83-dimensional vectors consisting of an 80-dimensional log Mel filterbank (FBANK) extracted by Kaldi\footnote{https://github.com/kaldi-asr/kaldi} and 3-dimensional pitch information.
We preprocessed the text data with Byte Pair Encoding (BPE) to split the sentences into subwords with SentencePiece \cite{kudo2018sentencepiece}.
A dictionary with a vocabulary of 8,000 words was shared between the source and target languages.
\par
To evaluate the performance, we aligned the outputs of a model for automatically segmented speeches by each segmentation method (section \ref{subsec:method}) to the reference text with an edit distance-based algorithm \cite{matusov2005evaluating}.
We then calculated WER for ASR and BLEU for MT and ST.

\renewcommand{\arraystretch}{1.1}

\begin{table}[t]
\centering
\caption{Model settings: {\dag}version 0.10.3}
\label{table:config}
\begin{tabular}{|l|clc|}
\hline
Settings (ESPnet{\dag} options)                                                            & \multicolumn{1}{c|}{ASR} & \multicolumn{1}{l|}{ST}  & \multicolumn{1}{l|}{MT} \\ \hline\hline
Epochs (epochs) & \multicolumn{1}{c|}{45}  & \multicolumn{2}{c|}{100}                           \\ \hline
Encoder layers (elayers)                                                          & \multicolumn{2}{c|}{12}                             & 6                       \\ \hline
Decoder layers (elayers)                                                          & \multicolumn{3}{c|}{6}                                                        \\ \hline
FNN dimensions (eunits, dunits)                                                      & \multicolumn{3}{c|}{2048}                                                     \\ \hline
Attention dimensions (adim)                                                          & \multicolumn{3}{c|}{256}                                                      \\ \hline
Attention heads (aheads)                                                       & \multicolumn{3}{c|}{4}                                                        \\ \hline
Mini-batch (batch-size)                                                           & \multicolumn{2}{c|}{64}                             & 96                      \\ \hline
Gradient accumulation (accum-grad)                                                             & \multicolumn{2}{c|}{2}                              & 1                       \\ \hline
Gradient clipping (grad-clip)                                                          & \multicolumn{3}{c|}{5}                                                        \\ \hline
Learning rate (transformer-lr)                                                          & \multicolumn{1}{c|}{5}   & \multicolumn{1}{l|}{2.5} & 1                       \\ \hline
\begin{tabular}[c]{@{}l@{}}Warmup\\  (transformer-warmup-steps)\end{tabular} & \multicolumn{3}{c|}{25000}                                                    \\ \hline
Label smoothing (lsm-weight)                                                           & \multicolumn{3}{c|}{0.1}                                                      \\ \hline
Attention dropout (dropout-rate)                                                       & \multicolumn{3}{c|}{0.1}                                                      \\ \hline
\end{tabular}
\end{table}

\subsection{ST systems} \label{subsec:st}
We used the Transformer implementation of ESPnet\footnote{https://github.com/espnet/espnet} \cite{watanabe2018espnet, inaguma-etal-2020-espnet} to build the ASR and MT models for the cascade ST and an ST model for the end-to-end ST.
For the ASR and ST models using acoustic features as input, we added a 2D-convolution layer before the transformer encoder layers.
The model settings are shown in Table \ref{table:config}.
We trained the ASR model with hybrid CTC/attention \cite{watanabe2017hybrid} that incorporates the CTC loss into the Transformer.
The weight on the CTC loss was set to 0.3.
The ST model parameters were initialized with the encoder of the ASR model.
The parameters of each model were stored for each epoch.
After training to the maximum number of epochs, we averaged the parameters of five epochs with the highest validation scores for evaluation.

\subsection{Segmentation methods} \label{subsec:method}
\subsubsection{Baseline 1: VAD}
As a baseline, we used WebRTC VAD, a GMM-based VAD.
We tried nine conditions in the range of \textit{Frame size}=\{10, 20, 30 ms\} and \textit{Aggressiveness}=\{1, 2, 3\}.

\subsubsection{Baseline 2: fixed-length}
Fixed-length is a length-based approach that splits speech at a pre-defined fixed length \cite{sinclair2014semi}.
Although over- and under-segmentations are likely to occur because the method does not take acoustic and linguistic clues into account, its advantage is that the segment length was kept constant.
We tried ten fixed-length settings with parameters ranging from 4 to 40 seconds at 4-second intervals.

\subsubsection{Speech segmentation optimization}
We performed speech segmentation using our proposed model described in section \ref{subsec:model}.
During training, two consecutive segments were concatenated, as described in section \ref{subsec:data}.
The model’s setting was almost identical as that of the ASR encoder shown in Table \ref{table:config}. 
However, since the speech segment concatenation roughly doubles the average length of the input, we set the number of mini-batches to 32 and the number of gradient accumulations to 4.
Fixed-length $T$ of the input during the inference was set to 20 seconds based on the best score of the validation data.

\subsubsection{Hybrid method}
We also performed the hybrid method with our model and VAD, as introduced in section \ref{subsubsec:hybrid}.
For the hybrid method, we used a WebRTC VAD with (frame size, aggressiveness) = (10 ms, 2) and set \textit{maxlen} to ten seconds.

\begin{figure}[t]
\centering
\includegraphics[keepaspectratio, width=8cm]{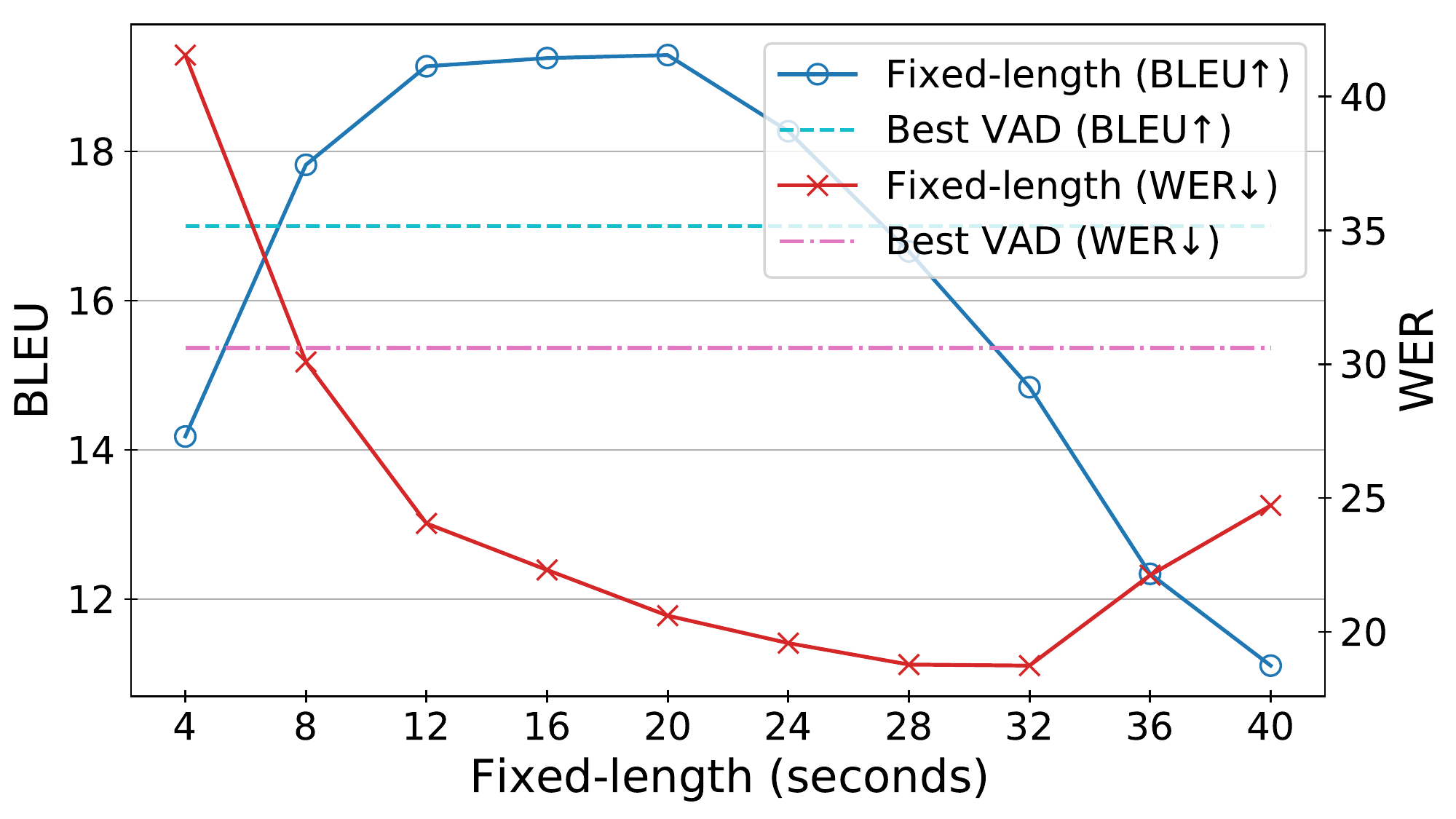}
\caption{Baseline scores on English-German cascade ST: Vertical axis indicates BLEU (left) and WER scores (right) for ASR and MT models.}
\label{fig:baseline}
\end{figure}

\section{Results and discussion}
\subsection{Baselines}
\label{subsec:results-baselines}
Figure \ref{fig:baseline} shows the ASR and MT results by cascade ST for each fixed-length.
For comparison, the scores for the setting with the highest BLEU of VAD are shown as straight lines (Best VAD).
With the fixed-length approach, WER and BLEU improved in proportion to the input length and deteriorated after reaching a certain length.
This result suggests that longer segments can prevent the degradation of the ASR and ST performances due to automatic segmentation.
On the other hand, we identified an upper limit to the segment length that the model successfully handled, which depends on its capability.
Therefore, automatic speech segmentation is important to prevent ST performance degradation.
In addition, the best VAD results were worse than the best fixed-length results (Best Fixed-length)\footnote{A similar trend was shown in a previous study \cite{DBLP:journals/corr/abs-2104-11710}.},
suggesting that over- and under-segmentation due to pause-based segmentation significantly reduced the ASR and ST performances.

\begin{table}[t]
\centering
\caption{Results measured in WER and BLEU for tst-COMMON on MuST-C English-German cascade and end-to-end ST.}
\label{table:ende}
\begin{tabular}{l|cc|c}
\hline
                        & \multicolumn{2}{c|}{Cascade ST} & End-to-end ST \\ \hline
                        & WER            & BLEU           & BLEU          \\ \hline
Oracle                  & 12.60          & 23.59          & 22.50         \\ \hline
Best VAD                & 30.59          & 17.02          & 16.40         \\
Best Fixed-length       & 20.60          & 19.29          & 17.96         \\
Our model               & 20.99          & 20.18          & 19.10         \\
~~+VAD hybrid & \textbf{19.06}   & \textbf{20.99}   & \textbf{19.87}   \\ \hline
\end{tabular}
\end{table}

\begin{table}[t]
\centering
\caption{Results measured in WER and BLEU for tst-COMMON on MuST-C English-Japanese cascade and end-to-end ST.}
\label{table:enja}
\begin{tabular}{l|cc|c}
\hline
                        & \multicolumn{2}{c|}{Cascade ST} & End-to-end ST \\ \hline
                        & WER            & BLEU           & BLEU          \\ \hline
Oracle                  & 9.30          & 12.50          & 10.60         \\ \hline
Best VAD                & 25.81          & 9.26          & 8.14         \\
Best Fixed-length       & 18.89          & 9.64          & 8.52         \\
Our model               & 16.21          & 9.71          & 8.77         \\
~~+VAD hybrid & \textbf{13.67}   & \textbf{10.60}   & \textbf{9.24}   \\ \hline
\end{tabular}
\end{table}

\subsection{Proposed method}
Table \ref{table:ende} shows the overall results of the English-German experiments.
Our model outperformed VAD and the fixed-length segmentations for both cascade and end-to-end STs.
Roughly speaking, there are improvements in +3 BLEU for the Best VAD and in +1 BLEU for the Best Fixed-length.
This suggests that our model can split the speech into segments that correspond to sentence-like units suitable for translation.
\par
In addition, the hybrid method with VAD significantly improved both cascade (-1.93 WER and +0.81 BLEU) and end-to-end STs (+0.77 BLEU).
We confirmed that a hybrid method with our model and VAD greatly improved the translation performance.
However, room for improvement remains compared to the oracle segments contained by the MuST-C corpus.
\par
Table \ref{table:enja} shows the overall results of the English-Japanese experiments.
They resemble those in English-German; our model outperformed the existing methods, and a hybrid with VAD achieved more significant improvements.
We conclude that the proposed method is also effective for distant language pairs.

\begin{table}[t]
\centering
\caption{Example of ASR and MT outputs with segmentation positions: $\blacksquare$ indicates segment boundaries.}
\label{table:case-exp}
\begin{tabularx}{\linewidth}{p{23.7mm}|X} \hline
\textbf{Oracle (ASR)} &
\textit{bonobos are together with chimpanzees you aposre living closest relative $\blacksquare$} \\
\textbf{Oracle (MT)} &
\textit{Bonobos sind zusammen mit Schimpansen, Sie leben am nachsten Verwandten. $\blacksquare$} \\ \hline
\textbf{Best VAD (ASR)} &
\textit{bonobos are $\blacksquare$ together with chimpanzees you aposre living closest relative that ...} \\
\textbf{Best VAD (MT)} &
\textit{Bonobos sind es. $\blacksquare$ Zusammen mit Schimpansen leben Sie im Verhaltnis zum ...} \\ \hline
\textbf{Our model (ASR)} &
\textit{bonobos are together with chimpanzees you aposre living closest relative $\blacksquare$} \\
\textbf{Our model (MT)} &
\textit{Bonobos sind zusammen mit Schimpansen, Sie leben am nachsten Verwandten. $\blacksquare$} \\ \hline
\end{tabularx}
\end{table}

\begin{figure}[t]
\centering
\includegraphics[keepaspectratio, width=7cm]{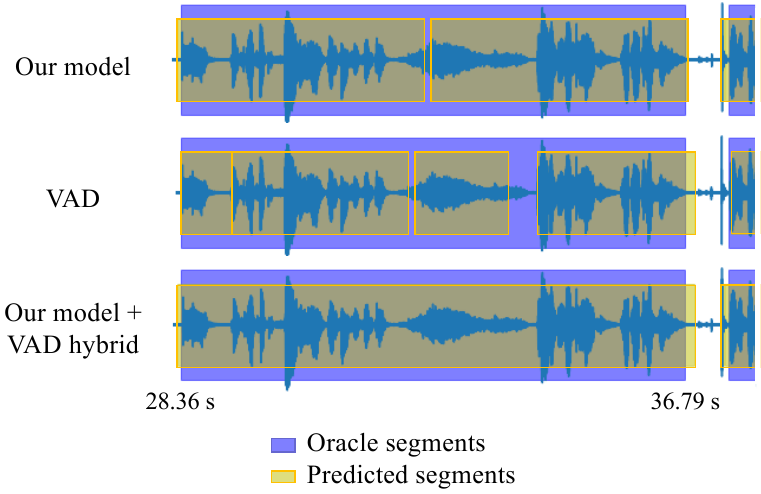}
\caption{Visualization of waveforms and segmentation positions}
\label{fig:case-wave}
\end{figure}

\subsection{Case study}
Table \ref{table:case-exp} shows an example of the ASR and MT outputs of the cascade ST using VAD and our segmentation model.
In VAD, more over- and under-segmentation occurred compared to the oracle segments.
In the example, the best VAD resulted in over-segmentation (``\textit{bonobos are $\blacksquare$ together}") and under-segmentation (``\textit{relative that ...}").
These errors caused differences in the oracle segments in the MT results.
On the other hand, our model split the speech at a boundary close to an oracle segment and obtained the same translation results.
\par
Figure \ref{fig:case-wave} shows an example of the waveforms and their segmentation positions.
The top two waveforms show that our model and VAD caused over-segmentation.
As the bottom waveform shows, hybrid decoding split the audio at a boundary near the oracle and
alleviated the problem by requiring an agreement between our model and the VAD.

\section{Conclusions}
We proposed a speech segmentation method based on a bilingual speech corpus.
Our method directly split speech into segments that correspond to sentence-like units to bridge the gap between training and inference.
Our experimental results showed the effectiveness of the proposed method compared to conventional segmentation methods on both cascade and end-to-end STs.
We also demonstrated that combining the predictions of our model and VAD further improved the translation performance.
\par
Future work will integrate a segmentation function into an end-to-end ST.
In this study, we treated segmentation as an independent process from ST, although it should be included in the \emph{end-to-end} ST.
We will investigate ways to integrate our proposed segmentation method into ST for streaming or online processing.
Future work will also further investigate different domains, other language pairs, and noisy environments and improve the segmentation model to reduce the need for VAD.

\section{Acknowledgements}
Part of this work was supported by JST SPRING Grant Number JPMJSP2140 and JSPS KAKENHI Grant Numbers JP21H05054 and JP21H03500.
\clearpage
\bibliographystyle{IEEEtran}

\bibliography{IS2022_ryo-fukuda_final}

\end{document}